\documentclass[conference]{IEEEtran}

\ifCLASSINFOpdf
  \usepackage[pdftex]{graphicx}
  \DeclareGraphicsExtensions{.pdf,.jpeg,.png}
\else
\fi

\usepackage{amsfonts, graphicx,amsmath,amsbsy,amssymb}

\newcommand{\ab}{{\mathbf a}}

\newcommand{\eb}{{\mathbf e}}

\newcommand{\rb}{{\mathbf r}}

\newcommand{\Kc}{{\mathcal K}}

\newcommand{\alphab}{\boldsymbol{\alpha}}

\newcommand{\omegab}{\boldsymbol{\omega}}

\newcommand{\sgn}{\mathrm{sgn}}

\begin{document}
%
\title{Multi-Scale Wavelet Domain Residual Learning\\for Limited-Angle CT Reconstruction}

\author{\IEEEauthorblockN{Jawook Gu}
\IEEEauthorblockA{
Korea Advanced Institute of Science and Technology\\
Daejeon, Korea\\
Email: jwisdom9299@kaist.ac.kr}
\and
\IEEEauthorblockN{Jong Chul Ye}
\IEEEauthorblockA{
Korea Advanced Institute of Science and Technology\\
Daejeon, Korea\\
Email: jong.ye@kaist.ac.kr}}



\maketitle

\begin{abstract}
Limited-angle computed tomography (CT) is often used in  clinical applications such as C-arm CT for interventional imaging.
However, CT images from limited angles suffers from heavy artifacts due to incomplete projection data.
Existing iterative methods require extensive calculations but can not deliver satisfactory results.
Based on the observation that the artifacts from limited angles have some directional  property and
 are globally distributed,
we propose a novel multi-scale wavelet domain residual learning architecture, which compensates for the artifacts.
Experiments have shown that the proposed method  effectively eliminates artifacts, thereby preserving edge and global structures of the image.
\end{abstract}


%
\IEEEpeerreviewmaketitle

\section{Introduction}
Computed tomography (CT) is widely used for clinical diagnosis.
In classical CT applications it is assumed that projection data are obtained over all angular ranges.
For example, an exact reconstruction requires a consecutive $180^{\circ}$  or 180$^{\circ}$+fan angle scans for parallel or fan beam reconstruction, respectively  \cite{tam1988reducing}.
For limited angle scans,  however, the projection data will cover an angular range of less than $180^{\circ}$.
A limited-angle scan is used because of the large object size \cite{wu2003tomographic}, large-pitch helical CT \cite{li2006exact}, and restricted scanning \cite{gao2007volumetric}.
In interventional imaging with C-arm CT, the limited angle acquisition is a commonly used protocol due to the hardware limitations.

Since limited angular scanning provides only a small subset of complete projection data, the use of a conventional filtered-back projection (FBP) algorithm  generally produces images with heavy directional artifacts.
Therefore, many studies have been proposed to compensate for artifacts in limited-angle CT images.
Several iterative reconstruction algorithms such as singular value decomposition, wavelet decomposition \cite{rantala2006wavelet},  and reprojection have been developed.
Inspired by the success of compressed sensing (CS) algorithm in a sparse-view CT \cite{candes2006robust}, many CS-based methods have also been proposed in limited-angle CT.
A representative CS-based algorithm is POCS with total variation minimization  \cite{sidky2008image}.
%
%
Although these algorithms are good in restoring the localized artifacts or disturbances,
these algorithms have not been successful in correcting globalized artifact patterns in the CT image from a limited angle acquisition.
In addition,  the CS-based methods require a large amount of iterative computation.
Therefore, CS-based  methods need to improve image quality and reconstruction time.


Recently, deep learning algorithms using convolutional neural network (CNN) show successful results in computer vision applications including image classification \cite{krizhevsky2012imagenet}, denoising \cite{zhang2016beyond} and segmentation \cite{ronneberger2015u}.
Some studies have tried to use the deep learning algorithms for image reconstruction problems. In the CT area,
Kang et al  \cite{kang2016deep} proposed a wavelet-domain deep learning network for denoising for  low-dose CT and showed the promising results by winning second place 
in the AAPM Low-Dose Grand Challenge.
Jin et al \cite{jin2016deep} and Han et al \cite{han2016deep} independently proposed multi-scale residual learning network for sparse view CT reconstruction.
Recently, the first applications of CNN appeared for limited angle tomography \cite{zhang2016image}.
However, the CNN in \cite{zhang2016image}  used only  three layers and has not been explicitly designed to correct for the directional
artifacts in the limited angle tomography.

By extending our prior work \cite{han2016deep}, here  we propose a novel multi-scale  wavelet-domain residual learning network for limited-angle CT reconstruction.
In particular, the network is designed in a directional wavelet transform domain to exploit the directional property of the limited angle artifacts.
In addition,   instead of directly learning the artifact-free image,  our network is designed as  a residual network that  estimates the artifact directly in the wavelet domain.
To take into account the globally distributed artifacts,  a multi-resolution network architecture was  used, inspired by  \cite{jin2016deep,han2016deep}. 
Once the wavelet domain residual is estimated,  artifact-free wavelet coefficients are estimated by subtracting the residuals, after which
the wavelet recomposition is performed to obtain the full resolution image.
Numerical results confirmed the proposed wavelet domain deep residual learning exceeds the existing methods in image quality and reconstruction time.

%
%

\section{Theory}

\subsection{Spectrum of limited angle artifacts}

\begin{figure}[!h]
\centering
\includegraphics[width=8cm]{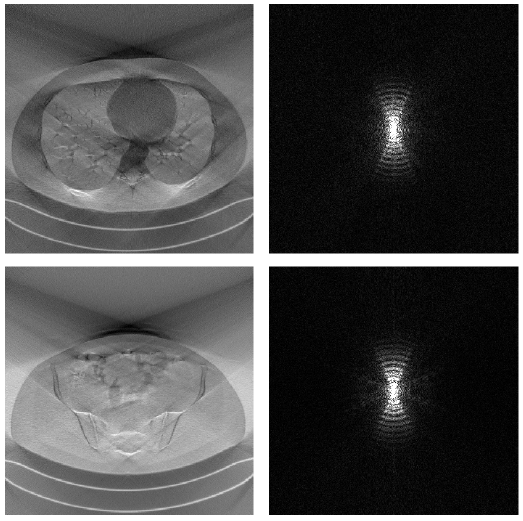}
\centerline{\mbox{(a)}\hspace{4cm}\mbox{(b)}} 
\caption{Spectrum analysis of limited angle artifacts. (a) The artifact images of $120^{\circ}$ limited angle CT and (b) their corresponding spectra.}
\label{fig:spectrum}
\end{figure}

 \begin{figure}[!h]
\centering
\includegraphics[width=8cm]{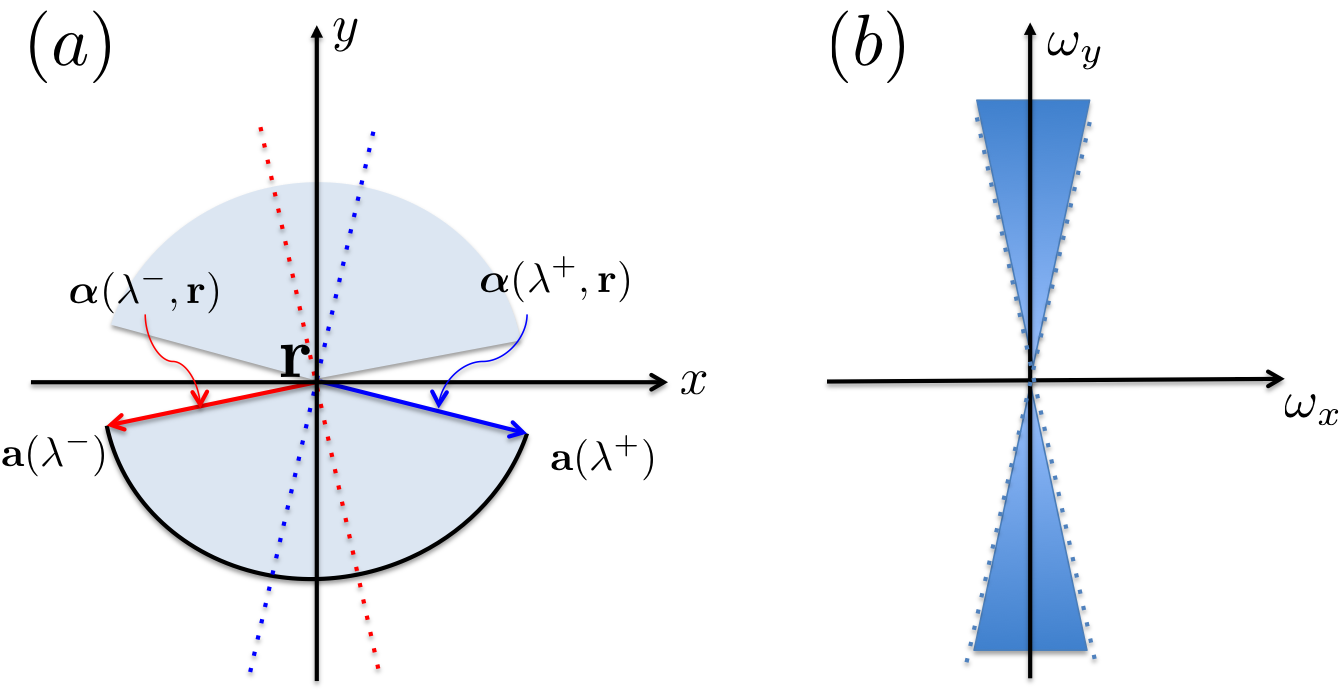} 
\caption{(a) Limited scanning between angle $[\lambda^-,\lambda^+]$ and the associated vectors. (b) The corresponding missing frequency region.}
\label{fig:angle}
\end{figure}

Figure \ref{fig:spectrum} shows spectral analysis for limited angle CT artifact images that correspond to the difference between the full view and limited view reconstruction.
Specifically, Fig.~\ref{fig:spectrum}(a) shows the $120^{\circ}$ limited angle artifact image corresponding to the  scanning trajectory  in Fig.~\ref{fig:angle}(a), 
and Fig.~\ref{fig:spectrum}(b) show their corresponding spectra.
Despite different image contents, the spectral components of the artifacts have similar directional characteristics since the spectral components of the artifacts are determined by the missing angle.

This phenomenon can be analyzed using the spectrum of limited-angle reconstruction.
The Fourier slice theorem can show the missing frequency bands especially for parallel beam geometry.
For general scanning geometry,  the Katsevich formula \cite{katsevich2004improved} can be used
to analyze this. 
Here,  we consider  2-D fanbeam scanning geometry  of objects within a field of view $\Omega$.
Then, for a given a filtering direction $\eb$ and the x-ray source angle between $\lambda^-$ and $\lambda^+$, 
the  Katsevich integral $\Kc(\rb,\eb,\lambda^-, \lambda^+) $ at $\rb\in \Omega$   can be represented as follows \cite{pack2005cone}:
\begin{eqnarray}
\Kc(\rb,\eb,\lambda^-, \lambda^+)   
&=& \frac{1}{(2\pi)^2} \int d\omegab \hat\mu(\omegab)  \sigma(\rb,\omegab) e^{j\omegab\cdot \rb} \label{eq:K}
\end{eqnarray}
Here, $\hat \mu(\omegab)$ is the Fourier transform of the image $\mu(\rb)$ and 
\begin{eqnarray*}
\sigma(\rb,\omegab) = \frac{1}{2} \sgn(\omegab\cdot \eb) [\sgn(\omegab \cdot \alphab(\lambda^-,\rb)) - \sgn(\omegab \cdot \alphab(\lambda^+, \rb))] \label{eq:sigma_fourier}
 \end{eqnarray*} 
 where $\alphab(\lambda,\rb)$ is the vector connecting the x-ray source at $\lambda$ and the reconstruction position $\rb$ (see Fig.~\ref{fig:angle}(a)).
 The filtering direction $\eb$ is often set to $ \alphab(\lambda^-,\rb)$.
 
Note that \eqref{eq:K} is identical to the inverse Fourier transform if $\sigma(\rb,\omegab)=1,\forall \omegab$ . This can be achieved in  a full  scanning geometry
because, for any $\rb \in \Omega$, we can find $\ab(\lambda^-)$ and $\ab(\lambda^+)$ such that
 $\alphab(\lambda^-,\rb)=- \alphab(\lambda^+,\rb)$.
However, with  limited angle scanning, there exist
frequency $\omegab$  such that $\sgn(\omegab \cdot \alphab(\lambda^-,\rb)) = \sgn(\omegab \cdot \alphab(\lambda^+, \rb))$,  providing    $\sigma(\rb,\omegab)=0$.
Given the limited scanning geometry in Fig.~\ref{fig:angle}(a), for   reconstructing the pixel on the isocenter,
the $\omegab$ vectors between the red and blue dashed lines results in  $\sgn(\omegab \cdot \alphab(\lambda^-,\rb)) = \sgn(\omegab \cdot \alphab(\lambda^+, \rb))$.
In fact, this corresponds to the missing frequency region described in Fig.~\ref{fig:angle}(b), resulting in an elongated artifact along y-direction.
The missing frequency region depends on $\rb$, which is why we have additional components in  Fig.~\ref{fig:spectrum}(b),  but we can still see that the general directional
characteristics of the artifacts are similar.



This suggests design guidelines for the deep network.
First, in  limited angle tomography, the artifacts, even if the artifact-free images are drastically different from each other, are similar to each other. This means that learning the artifact is easier than learning artifact-free original images. Secondly, the artifacts have a strong directional characteristic so that the learning on the directional wavelet domain is preferred to the original image domain.
Finally, as shown in Fig.~\ref{fig:spectrum}(a), the artifacts are distributed globally, so the network should have large receptive field.
Inspired by these observations, we proposed a multi-scale  wavelet domain  residual learning network  for limited-angle tomography.


\subsection{Directional wavelet transform using contourlets}

Specifically,  CT reconstruction images are first decomposed using non-decimated redundant directional wavelet transform, called 
contour transformation \cite{da2006nonsubsampled}. The  contourlet transformation performs a multiscale directional decomposition of an image. Here, the image is divided into high frequency components
and low-frequency components, and the directional filter banks are applied to the divided frequency components.
More specifically, when $k$ denotes the number of multiscale decomposition stage, the number of channels in directional decomposition is $2^k$.
We used four stage of multiscale decomposition.
The total number of channels in a wavelet domain is therefore $2^0 + 2^1 + 2^2 + 2^3 = 15$.

\begin{figure*}[t]
\centering
\includegraphics[width=12cm]{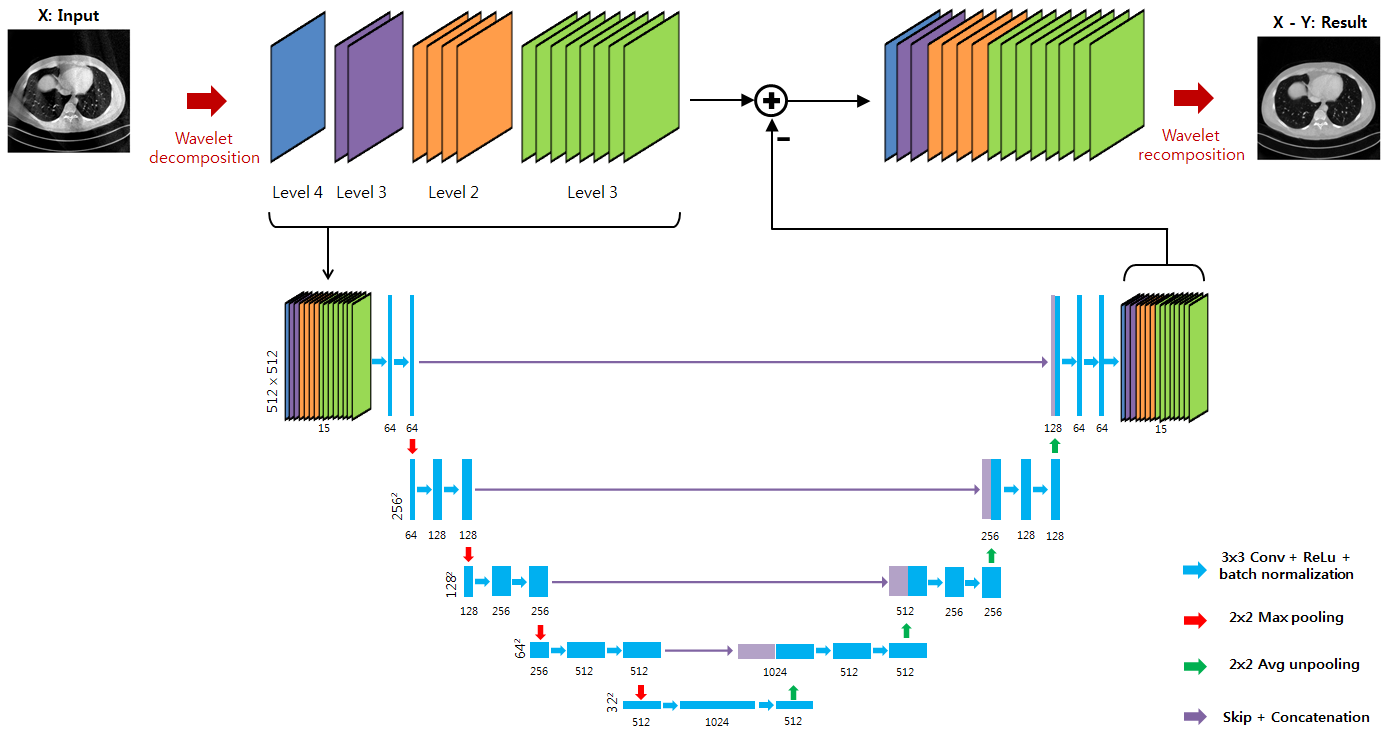} 
\caption{The proposed multi-scale wavelet domain deep residual learning architecture for limited-angle CT reconstruction.}
\label{fig_architecture}
\end{figure*}

\subsection{Deep residual network architecture}

Then, we developed a wavelet domain deep residual network by modifying the U-Net \cite{ronneberger2015u}.
The CNN architecture is shown in Fig. \ref{fig_architecture}.
A basic operation, a blue arrow in Fig. \ref{fig_architecture}, consists of $3 \times 3$ convolutions followed by a rectified linear unit (ReLU) and batch normalization.
The U-Net architecture consists of a contracting path and an expansive path.
In the contracting path, $2 \times 2$ max pooling operation follows after two basic operations.
After $2 \times 2$ max pooling operation, a number of channels is doubled.
In contrast,  $2 \times 2$ average unpooling operation is used in the expansive path  instead of 
$2 \times 2$ max pooling operation.
A {skip and concatenation operation}, represented  by violet arrows in Fig. \ref{fig_architecture}, directly concatenates  the results in the contracting path and the results in the expansive path. 
This multi-scale deep network structure in U-net is shown to  increase the receptive field so that it can effectively capture the globally distributed artifacts as the limited angle artifacts \cite{han2016deep}.

Once the artifact is estimated using contourlet domain U-net, then artifact-free wavelet coefficients are estimated by subtracting the residuals, after which
the wavelet recomposition is performed to obtain the full resolution image.

\section{Method}

\subsection{Data Set}

We used nine sets of real projection data from AAPM Low-Dose CT Grand Challenge.
The provided data sets are acquired in helical CT, so we rebinned the projection data from the helical CT to $360^{\circ}$ angular scan fan-beam CT.
The $512 \times 512$ artifact free images are reconstructed using full-angle fan-beam projection data. 
In limited angle experiments, only $120^{\circ}$ or $150^{\circ}$ angular scan were used to reconstruct limited-angle CT images.
Every reconstruction process used traditional FBP algorithm.
Eight sets out of nine data sets were used for network training.
The remaining set was used for validation.

\subsection{Network training}

We developed the proposed network using MatConvNet
toolbox in MATLAB R2015a environment \cite{vedaldi2015matconvnet}. GTX 1080
graphics processing unit and Intel Core i7-4790 central processing
unit were used for research. 
The network is trained by 256$\times$ 256 size randomly selected patches. Then, the learned filtered are used to process 512$\times$512 images at the test phase.
Training time lasted about 24 hours.
Stochastic gradient reduction was used to train the network. The number of epochs was 150. The initial learning rate was $10^{-3}$, which gradually dropped to $10^{-5}$. The regularization parameter was $10^{-4}$.

\subsection{Other baseline network architectures}

\begin{figure}[!h]
\centering
\includegraphics[width=3.5 in]{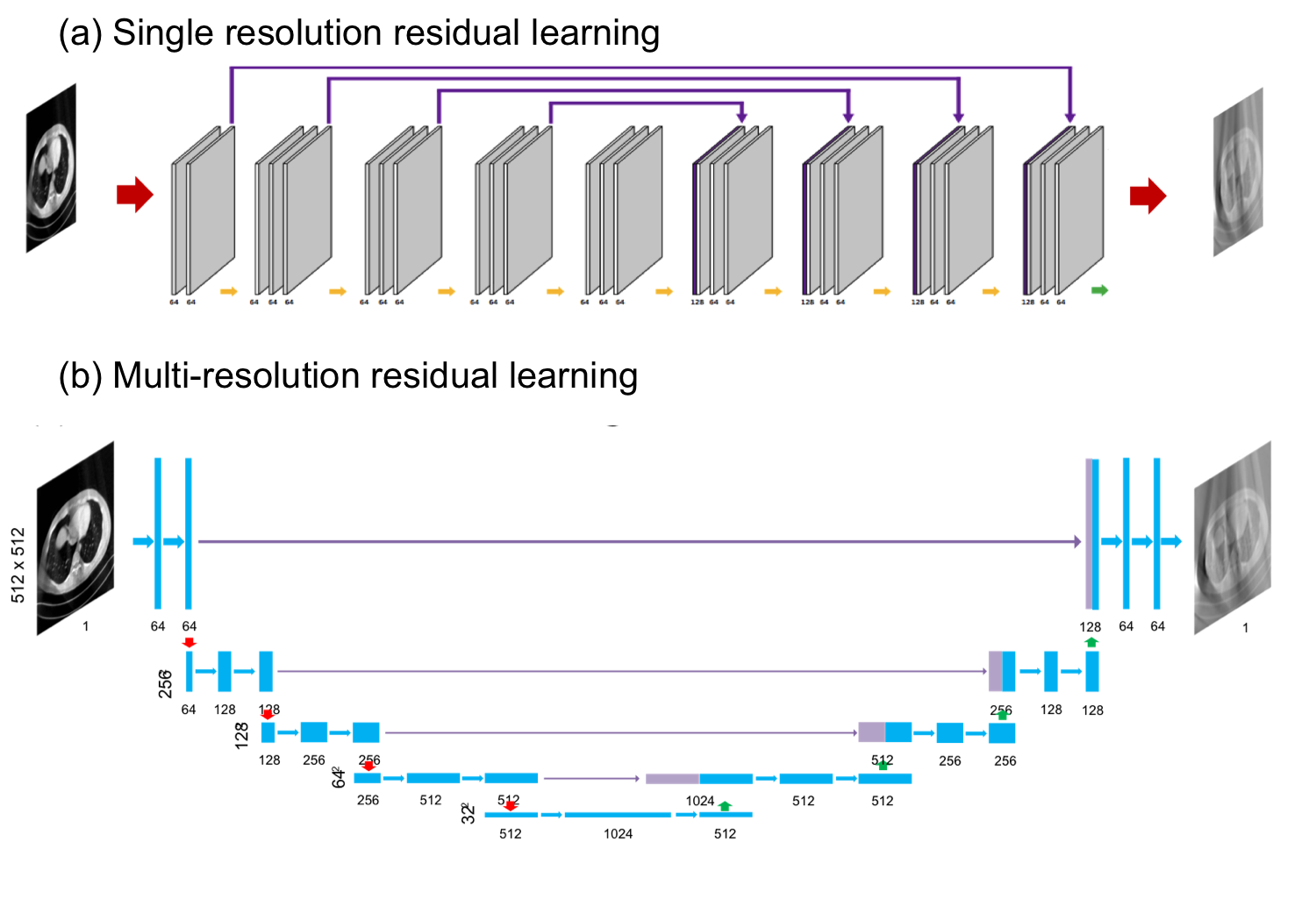} 
\caption{Architecture of two baseline networks.}
\label{fig_ref_networks}
\end{figure}

In order to investigate the optimal nature of the proposed network, 
we compared the proposed method with two different neural networks.
First network is a  single-resolution residual learning CNN network.
The first network eliminates the artifact in the image domain and uses single-resolution network architecture in Fig. \ref{fig_ref_networks}(a).
The second network is a multi-resolution residual learning CNN network as shown in Fig. \ref{fig_ref_networks}(b).
The second network directly uses the CT image instead of decomposed wavelet coefficients. 

\section{Results}



\begin{figure}[t]
\centering
\includegraphics[width=10cm]{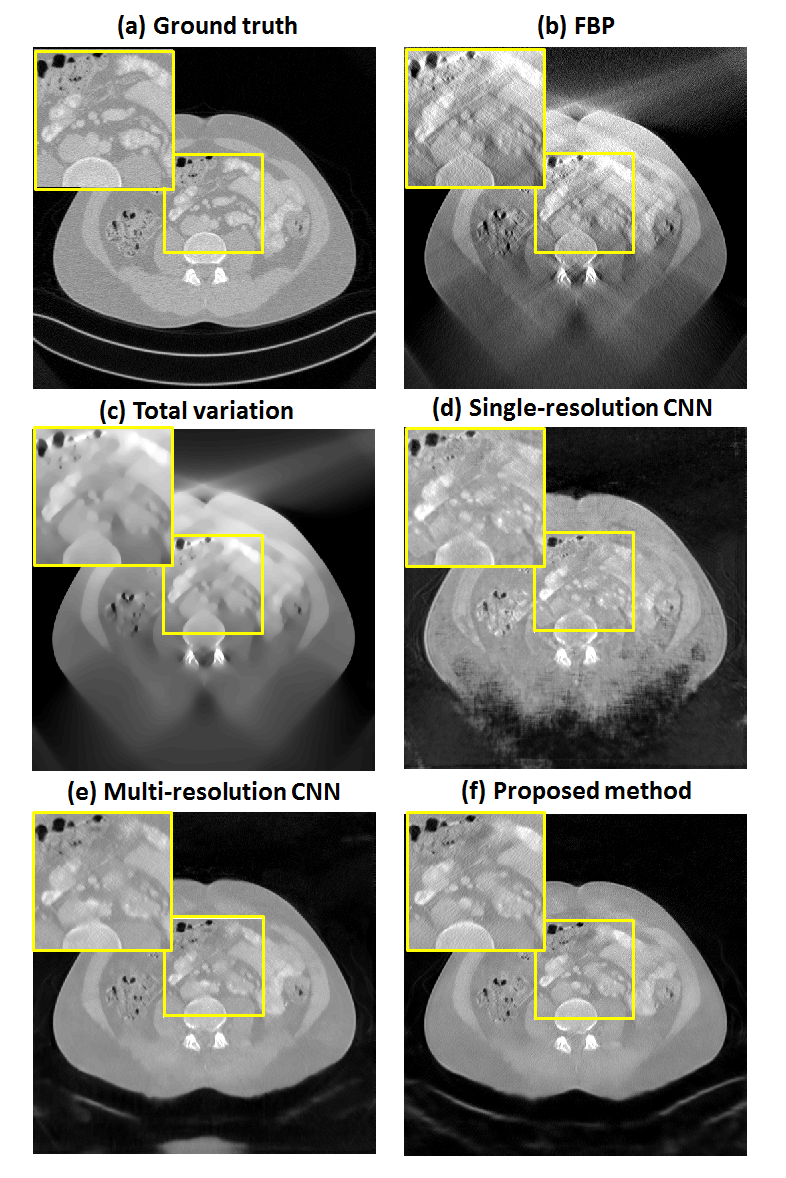} 
\caption{Reconstruction results by various methods for 120$^\circ$ scanning geometry.}
\label{fig_results120}
\end{figure}

The reconstruction results by FBP, TV,  baseline network  and the proposed method for 120$^\circ$ scanning are shown in Fig. \ref{fig_results120}.
 The single resolution CNN performs better than the TV approaches, but the internal and outer area are incorrectly recovered.
 On the other hand, the multi-resolution CNN  is significantly better than the single resolution CNN. However, there are remaining ghost artifacts beneath the patient bed,
 and the internal structures are reconstructed blurry.  The proposed multi-resolution wavelet domain residual network allowed for a  clear reconstruction
 and  effectively compensated the artifacts of CT compared to the other methods.
Comparing the entire image, the global artifacts are removed and the boundary of body  is clearly recovered in the proposed method.
The enlarged images, the yellow boxes in Fig. \ref{fig_results120},  show that the proposed method eliminates small local artifacts and reconstructs the detail structures.
It is remarkable that the proposed method successfully preserves the detail edge structures while compensating for heavy directional and blurring artifacts. 
Moreover, the computation time of the proposed method is about 0.34 sec/slice. It is about 10 times faster than the TV method which takes $3\sim4$ sec/slice as a computation time.

\begin{figure}[t]
\centering
\includegraphics[width=9cm]{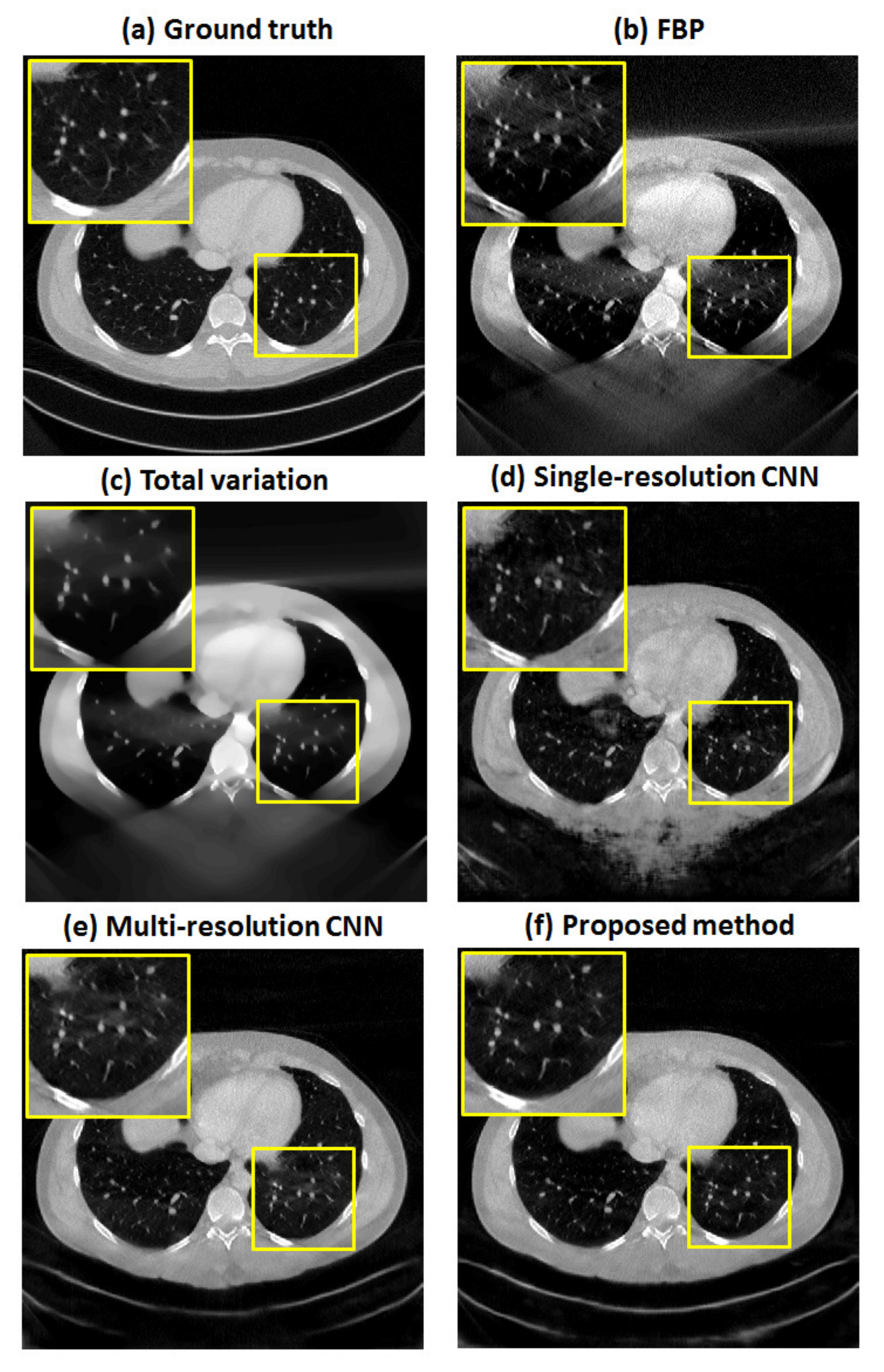} 
\caption{Reconstruction results by various methods for 150$^\circ$ scanning geometry.}
\label{fig_results150}
\end{figure}

For the 150$^\circ$ scanning experiments in  Fig. \ref{fig_results150},  similar reconstruction behaviours can be observed,
although the  visual difference between the multi-resolution CNN and multi-resolution wavelet domain CNN is not as significant as for the 120$^\circ$ cases.
The enlarged images, the yellow boxes in Fig. \ref{fig_results150},  show that wavelet domain multi-resolution CNN has less artifacts than the image domain
multi-resolution CNN.
For small missing angles,  however, both  the image domain and the wavelet domain multi-resolution residual network exhibited excellent subjective performances.

In Table ~\ref{tab_methods} and Table \ref{tab_methods2}, numerical analysis of image quality for different method, which are average values
from 488 slices of images,  are arranged.
All deep learning approaches showed better performance than the TV method.
However, the proposed method achieved the highest value in peak signal-to-noise ratio (PSNR) and structural similarity (SSIM),  and the lowest value in normalized root mean square error (NRMSE). The results of numerical analysis and reconstruction images confirmed that the proposed method exceeded the previous methods for CT reconstructions with a limited angle.

\begin{table}[t!] 
\caption{Comparison of various methods  for $120^{\circ}$ limited angle CT compensation.}
\label{tab_methods}
\begin{center}
\begin{tabular}{c|c|c|c|c|c}
\hline\hline
&  FBP &  TV  &Fig.~\ref{fig_ref_networks}(a) &Fig.~\ref{fig_ref_networks}(b)  & Proposed\\
\hline
PSNR [dB] &
18.4696 &
17.3981 &
23.6735 &
25.8605  &
26.5947 \\
NRMSE &
0.1284 &
0.1466 &
0.0798& 
0.0627&
0.0573 \\
SSIM &
0.4313 &
0.4221&
0.5164&
0.6762&
0.7173 \\
\hline\hline
\end{tabular}
\end{center}
\end{table}

\begin{table}[t!] 
\caption{Comparison of various methods  for $150^{\circ}$ limited angle CT compensation.}
\label{tab_methods2}
\begin{center}
\begin{tabular}{c|c|c|c|c|c}
\hline\hline
&  FBP &  TV  &Fig.~\ref{fig_ref_networks}(a) &Fig.~\ref{fig_ref_networks}(b)  & Proposed\\
\hline
PSNR [dB] &
21.4167 &
20.3215 &
26.2334 &
28.0536 &
28.7589 \\
NRMSE &
0.0980 &
0.1135 &
0.0559 &
0.0452 &
0.0429 \\
SSIM &
0.5089 &
0.4780 &
0.6304  &
0.7381 &
0.7453 \\
\hline\hline
\end{tabular}
\end{center}
\end{table}

\section{Conclusion}

In this work, we proposed a novel deep residual learning network for CT reconstruction with a limited angle projection data. The proposed network used U-network structure to increase the receptive field. Since the typical artifacts in the limited CT show globalized patterns like directional streaks and blurs, the network was trained on a directional wavelet transform domain.
Compared to FBP and TV methods, the proposed network shows excellent results.


\section*{Acknowledgment}

The authors would like to thanks Dr. Cynthia McCollough,  the Mayo Clinic, the American Association of Physicists in Medicine (AAPM), and grant EB01705 and EB01785 from the National
Institute of Biomedical Imaging and Bioengineering for providing the Low-Dose CT Grand Challenge data set.
This work is supported by Korea Science and Engineering Foundation, Grant number
NRF-2016R1A2B3008104.

 

%






\end{document}